\documentclass{article}

\usepackage[preprint]{corl_2024} 
\usepackage{graphicx}
\usepackage{booktabs}
\usepackage{multirow}
\usepackage{multicol}
\usepackage{makecell}
\usepackage{amsmath}
\usepackage{xcolor}
\usepackage{amssymb}
\usepackage{tikz}
\usepackage[page,header]{appendix}
\usepackage{titletoc}
\usepackage{enumitem}

\title{\textit{Tell Me Where You Are}: \\ Multimodal LLMs Meet Place Recognition\\
  {\tt\small\url{https://ai4ce.github.io/LLM4VPR/}}}

\author{
  Zonglin Lyu, Juexiao Zhang, Mingxuan Lu, Yiming Li, Chen Feng\\
  New York University \\
  \texttt{\{zl3958, jz4725, ml8465, yimingli, cfeng\}@nyu.edu} \\
}

\begin{document}
\maketitle

\vspace{-5mm}
\begin{abstract}
Large language models (LLMs) exhibit a variety of promising capabilities in robotics, including long-horizon planning and commonsense reasoning. However, their performance in place recognition is still underexplored. In this work, we introduce multimodal LLMs (MLLMs) to visual place recognition (VPR), where a robot must localize itself using visual observations. Our key design is to use \textit{vision-based retrieval} to propose several candidates and then leverage \textit{language-based reasoning} to carefully inspect each candidate for a final decision. Specifically, we leverage the robust visual features produced by off-the-shelf vision foundation models (VFMs) to obtain several candidate locations. We then prompt an MLLM to describe the differences between the current observation and each candidate in a pairwise manner, and reason about the best candidate based on these descriptions.  Our results on three datasets demonstrate that integrating the \textit{general-purpose visual features} from VFMs with the \textit{reasoning capabilities} of MLLMs already provides an effective place recognition solution, \textit{without any VPR-specific supervised training}. We believe our work can inspire new possibilities for applying and designing foundation models, \textit{i.e.}, VFMs, LLMs, and MLLMs, to enhance the localization and navigation of mobile robots.

\end{abstract}

\vspace{-3mm}
\begin{figure*}[h]
    \centering
    \includegraphics[width=0.9\linewidth]{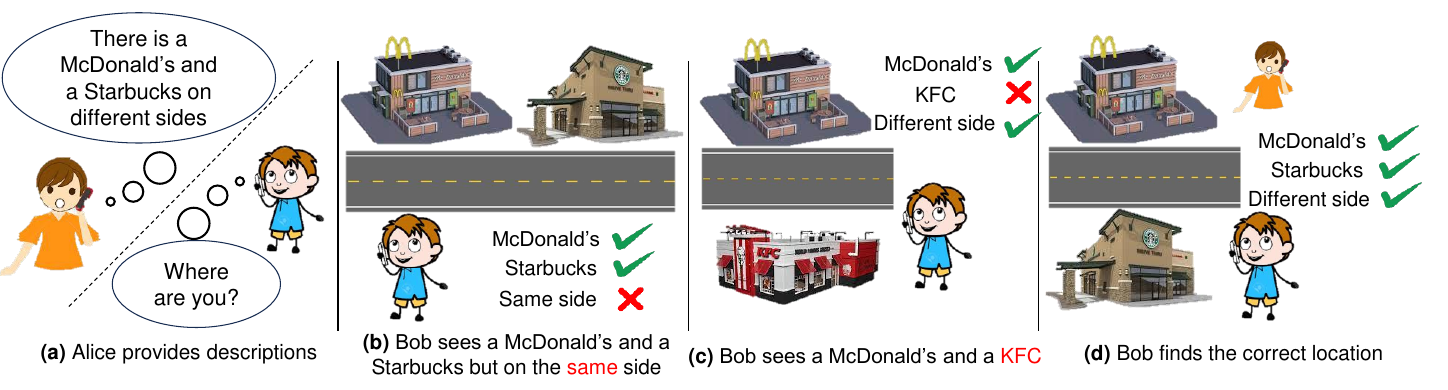}
    \caption{\textbf{Vision and language meet place recognition.} Alice gives Bob verbal descriptions of her surroundings (a). Bob compares his visual observations with Alice's descriptions (b)-(d) and reasons about their accuracy, confirming (d) as the correct place.}
    \label{fig:teaser}
\end{figure*}
\vspace{-3mm}

\section{Introduction}
\label{sec:intro}
Visual place recognition (VPR), the task of accurately identifying a previously visited location based on visual input, is a longstanding challenge in robotics~\cite{lowry2015visual}. Prior methods typically approach VPR as a visual representation learning problem, focusing on improving the robustness of visual features against irrelevant variations like lighting, weather, and transient objects. Inspired by the strong reasoning capabilities of large language models (LLMs), we explore how LLMs can enhance VPR performance. Intuitively, humans often use verbal descriptions alongside visual input to recognize their current location. Imagine a \textit{Bob and Alice} scenario shown in Fig.~\ref{fig:teaser}. Bob is trying to find Alice and needs to verify his location using Alice's verbal description: ``There is a McDonald's and a Starbucks on different sides." Alice provides no navigation commands, only descriptions of her surroundings for Bob to recognize the correct place, aligning with the VPR task definition. Bob compares his visual observations with Alice's cues and reasons about the correctness of the location: mismatched store positions indicate he is in the wrong place, while matching observations confirm he is in the right place. It seems intuitive to integrate language into VPR based on the aforementioned example.  \textit{Yet it remains unexplored how and to what extent LLMs can improve place recognition.}

\begin{figure*}[t]
    \centering
    \includegraphics[width=\linewidth]{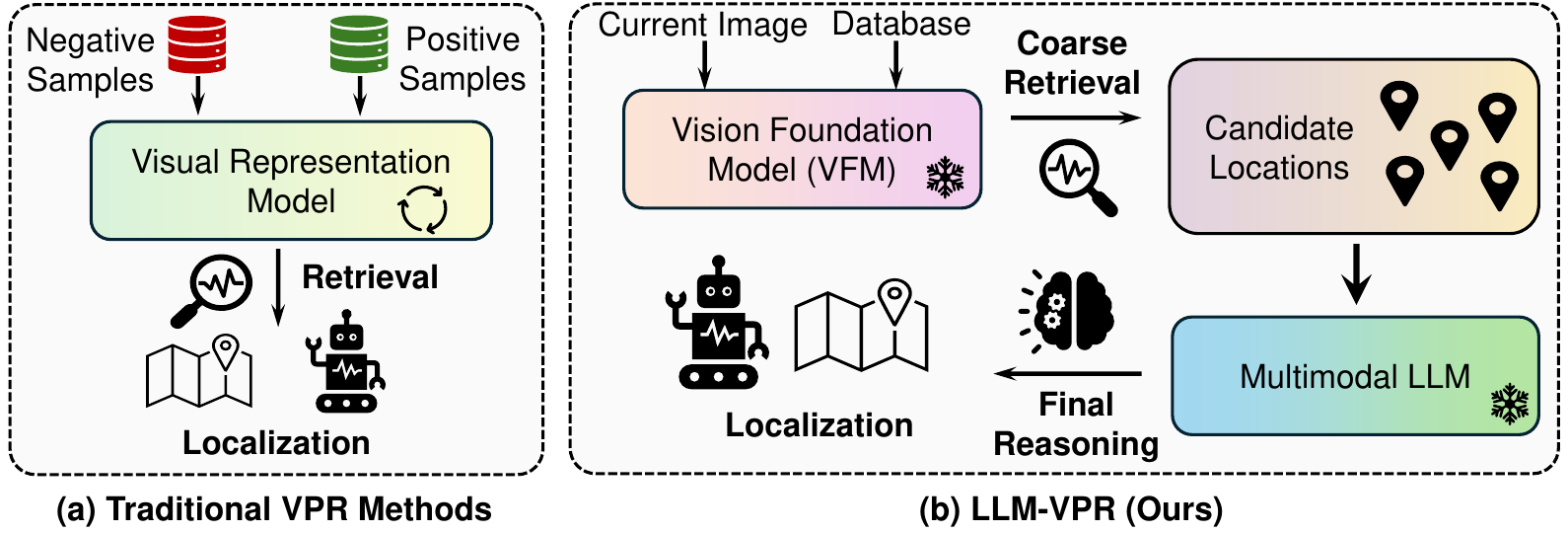}
    \caption{\textbf{Comparison between traditional VPR methods and our LLM-VPR.} We build a VPR solution based on off-the-shelf foundation models and do not need VPR-specific supervised training. Meanwhile, we leverage language-based reasoning to further refine the localization precision.}
    \label{fig:comparison}
    \vspace{-4mm}
\end{figure*}

Vision and language exhibit distinct properties. Visual observations offer abundant spatial details, including geometric and photometric cues. In contrast, \textit{language provides abstract information that facilitates spatial reasoning, such as identifying landmarks and understanding contextual relationships within an environment.} To leverage the strengths of both modalities, we propose a vision-to-language (coarse-to-fine) framework, in which vision-based retrieval generates several candidate locations, followed by language-based reasoning for finer selection. More specifically, we utilize an off-the-shelf vision foundation model (VFM), DINOv2~\cite{oquab2023dinov2}, to extract robust visual features from RGB input. These robust features enable coarse retrieval, proposing several candidate images. Subsequently, we employ an off-the-shelf multimodal LLM, GPT-4V~\cite{Achiam2023GPT4TR}, to perform a finer selection from these candidates. Nevertheless, fully harnessing the potential of multimodal LLMs without task-specific training to find the best candidate remains a challenge.

When asked to match two visual observations, humans compare both images by \textit{looking back and forth}: they identify potential matching landmarks, examine each one closely, and seek contextual cues to draw conclusions~\cite{chun2000contextual}. This hints at a \textit{comparing-then-reasoning} design, \textit{i.e.}, we leverage multimodal LLMs to describe the \textit{delta} of each image pair and then use all the textual descriptions for the final reasoning stage to determine the best candidate. Moreover, without task-specific fine-tuning, it is challenging for multimodal LLMs to identify VPR-relevant details, such as buildings, and VPR-irrelevant details, such as lighting, which can be easily done with human knowledge. Hence, we inject human knowledge into multimodal LLMs via VPR-specific prompts. Consequently, we build a VPR system with two off-the-shelf foundation models, without any additional supervised training used in previous VPR methods~\cite{Arandjelovi2015NetVLADCA,Kim2017LearnedCF,Zhang2021VectorOL,Alibey2023MixVPRFM,izquierdo2023optimal}, as shown in~Fig.~\ref{fig:comparison}.

\textbf{Statement of contributions.} We propose \textbf{LLM-VPR}, a framework integrating language and vision for robotic place recognition. 
We make the following contributions: (1) We demonstrate that integrating the \textit{general-purpose visual features} from VFMs with the \textit{reasoning capabilities} of MLLMs already provides an effective VPR solution, \textit{without any VPR-specific supervised training}. In other words, we leverage two foundation models to achieve zero-shot place recognition. (2) We propose a vision-to-language (coarse-to-fine) framework to fully exploit the advantages of two modalities in VPR. (3) We propose a \textit{comparing-then-reasoning} framework to facilitate spatial reasoning with MLLMs, achieving fine-grained place recognition. (4) We
evaluate \textbf{LLM-VPR} in three datasets. Quantitative and qualitative results indicate that our method outperforms vision-only solutions and performs comparably to supervised methods without training overhead.

\section{Related Works}
\textbf{Visual Place Recognition (VPR).} Most VPR methods consider only the vision modality and the task is formulated as image retrieval. Traditionally, VPR involves aggregating hand-crafted features~\cite{Hunt2009SURFSR,Lowe1999ObjectRF} into a global descriptor~\cite{Jgou2010AggregatingLD,Arandjelovi2013AllAV}. Recent learning-based methods take advantage of deep features~\cite{He2015DeepRL,dosovitskiy2021an}. For retrieval methods, learning-based VPR techniques fall into two folds: global retrieval and local reranking. Global retrieval methods~\cite{Arandjelovi2015NetVLADCA,Alibey2023MixVPRFM,Zhang2021VectorOL,Kim2017LearnedCF} aggregate latent features extracted by Deep Neural Networks into a global descriptor. Reranking methods~\cite{Hausler2021PatchNetVLADMF,Wang2022TransVPRTP,Zhu2023R2FU} coarsely retrieve the top-K images with global retrievals and conduct a detailed comparison of local patch features, sacrificing efficiency to obtain better quality. 
Some works enhance VPR with different inputs such as video VPR~\cite{Garg2021SeqNetLD,Garg2021SeqMatchNetCL}, collaborative VPR~\cite{Li2023CollaborativeVP}, and lidar-based VPR~\cite{uy2018pointnetvlad,xia2021soe,hou2022hitpr}.
Recently, language models have been introduced to conventionally pure vision tasks. However, to the best of our knowledge, it remains unexplored how and to what extent language can benefit VPR.

\textbf{Vision Foundation Models in VPR.} VFMs refer to models trained with a wide range of images and their learned features demonstrate strong generalization capability. In terms of training methods, VFMs fall into two folds: supervised and unsupervised. Supervised vision foundation models~\cite{dosovitskiy2021an,He2015DeepRL,liu2022convnet, radford2021learning} are usually trained in a large-scale dataset with labels such as ImageNet1K~\cite{Deng2009ImageNetAL}. Such models can be transferred to different datasets/tasks by changing their output heads. Most VPR methods~\cite{Arandjelovi2015NetVLADCA,Alibey2023MixVPRFM,Wang2022TransVPRTP,Zhu2023R2FU} initialize model weights pretrained in such a way, which requires additional fine-tuning. Due to high annotation costs in supervised methods, self-supervised techniques are proposed so that models can learn the pattern of images without any annotations, and thus models can potentially learn from an incredible amount of unlabelled data~\cite{oquab2023dinov2,he2020momentum,chen2020improved,caron2020unsupervised,chen2020simple,caron2021emerging}. DINO~\cite{caron2021emerging,oquab2023dinov2} can produce robust and informative features in VPR task even without fine-tuning~\cite{AnyLoc,izquierdo2023optimal}, achieving training-free general-purpose solution for VPR. Our method leverages DINO as a vision-based coarse retriever and further improves its performance with the help of human language.

\textbf{Languages in VPR.} There are a few works at the intersection of languages and VPR. TextPlace~\cite{hong2019textplace} highlights the importance of text appearing in scenes for the VPR task, but it specifically requires text to be present in the scene. FM-Loc~\cite{Mirjalili2023FMLocUF} employs multimodal LLMs in specific cases of place recognition, requiring the scene to be an indoor environment that can be divided into different categories. The solutions presented in these methods are limited to specific VPR scenarios, whereas our method offers a general-purpose framework applicable to a wide range of scenarios.

\textbf{Multimodal LLMs in Robotics.} Introducing LLMs to robotics is recently emerged as an active research area~\cite{robotGPT, driess2023palme, saycan2022arxiv, codeaspolicies2022, huang2023voxposer, jiang2023vima}. There are several lines of works exploring the advantages of language in other relevant vision-based robotics tasks, such as autonomous driving~\cite{chen2023driving,jin2023adapt,Xu2023DriveGPT4IE}, navigation~\cite{krantz2023iterative,yang2023behavioral, dai2023think} and robot-human interactions~\cite{knowno2023}. They generally ask for high-level actions to control the vehicles or robots which take advantage of MLLMs strength in semantic understanding and reasoning. In contrast, VPR often requires highly detailed low-level descriptions of images, which is reportedly nontrivial for MLLMs~\cite{tong2024eyes}. However, the study of multimodal LLMs in visual place recognition is still underexplored. In this work, we aim to bridge this gap, potentially leading to more robust and versatile applications of multimodal LLMs in autonomous robots, including but not limited to topological localization and spatial reasoning.

\section{Preliminaries}
\label{sec:retrieve}
\textbf{VPR as Image Retrieval.} 
Given a query image $I_q$, the VPR task requires finding the images capturing the same location as $I_q$ from a database of previously visited images $\{I_r\}_{r=1}^{N}$. This is generally considered as retrieving the correct images from the candidate database and hence is evaluated with recall rates.
Specifically, suppose a feature extractor $F$ maps images to global descriptors, which are considered as the image descriptors.
Then the similarity scores of a query image $I_q$ and a reference image $I_r$ are given by the cosine similarities between their corresponding global descriptors. The top-K candidates are references with the highest K similarity scores. The R@K metric evaluates the percentage of correctly retrieved images in the top-K selected candidates.

\textbf{Feature Aggregation.} Modern deep learning models usually do not directly map images to global descriptors, but rather to convolutional feature maps or sequences of patch features from ViT.
Therefore, various aggregation methods have been proposed to obtain informative image descriptors for VPR. Notably, Generalized Mean Pooling (GeM)~\cite{Radenovic2018gem} is a training-free aggregator that achieves lightweight yet strongest performance in VPR. 
Our method uses GeM to aggregate features obtained from pretrained DINOv2.
It is computed as: $X_{GeM} = \frac{1}{N}\left(\sum_i^N X_i^p\right)^{\frac1p}$, where $X_i$ represents the output feature of a patch of an image and $N$ represents the number of patches.

\begin{figure*}[t]
    \centering
    \includegraphics[width=1\linewidth]{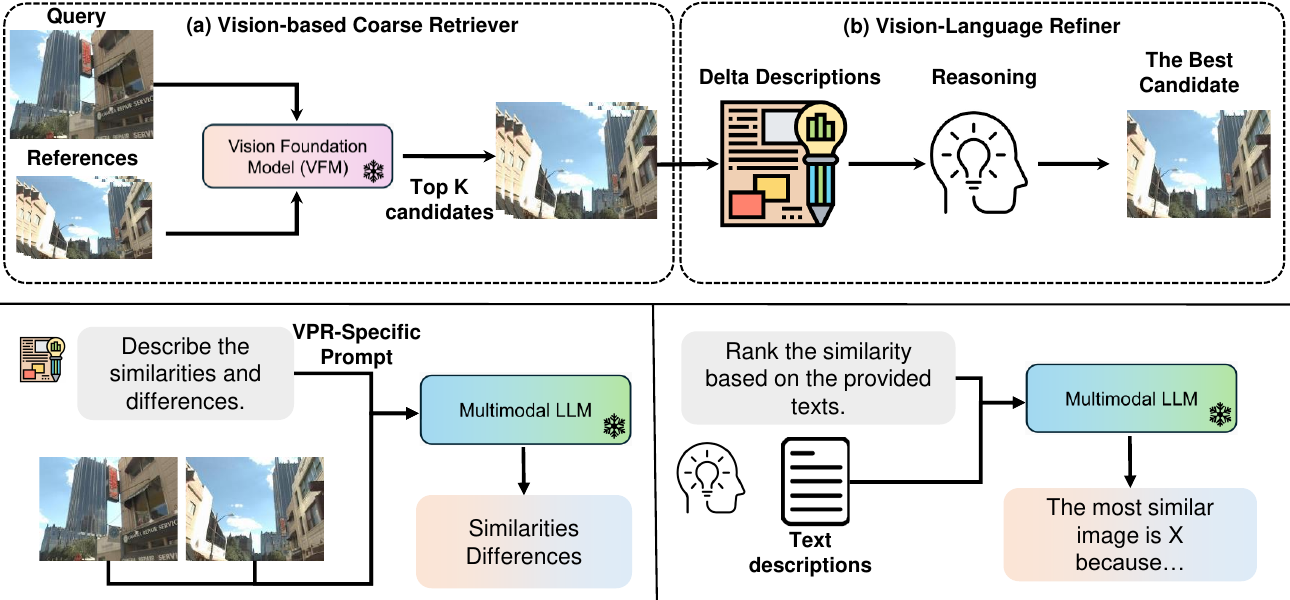}
    \caption{\textbf{Overview of LLM-VPR. (a) Vision-based Coarse Retriever. (b) Vision-Language Refiner.} We first coarsely retrieve top-10 candidates via \texttt{[CLS]} token or GeM aggregated descriptor of DINOv2~\cite{oquab2023dinov2} features. Then we construct ten query-candidate pairs and feed them one by one to the Vision-Language Refiner to describe and reason.}
    \label{fig:pipeline}
\end{figure*}

\section{Methodology}

\subsection{Design Rationale of LLM-VPR} 

Our method introduces MLLMs to VPR based on the following insights:

\begin{itemize}[leftmargin=1.3em]
    \item Vision and language exhibit distinct yet complementing attributes. Vision provides abundant details such as geometric and photometric cues, while language generates conceptual information and relations that aid spatial reasoning.
    \item Vision foundation models offer high-quality visual features that allow coarse retrieval to efficiently filter out apparently irrelevant candidates.
    \item Multimodal LLMs can extract detailed descriptions of the \textit{delta} between pairs of images via comparison and enable more sophisticated reasoning of their geographical relationships.
\end{itemize} 
Therefore, we propose a vision-to-language (coarse-to-fine) pipeline consisting of a \textit{vision-based coarse retriever} and a \textit{vision-language refiner}. The retriever selects the top-K candidates for a given query image, and the refiner generates a text description for each query-candidate pair and then evaluates all text descriptions to determine the similarity ranks between the query and each candidate. The entire pipeline is shown in Fig.~\ref{fig:pipeline}.

\subsection{Vision-based Coarse Retriever} The vision-based coarse retriever maps images to global descriptors and selects the top-K candidates based on the cosine similarity between the query and each candidate.
DINOv2~\cite{oquab2023dinov2} is employed as the vision feature extractor because its training is self-supervised by a large amount of task-agnostic vision data and exhibits strong VPR performance in previous work~\cite{AnyLoc}. Besides, DINOv2 is trained with both the global image-level loss and the local token-level losses, which produce more robust visual features for the VPR task. There are two approaches to retrieving the top-K candidates: directly using the \texttt{[CLS]} token or applying GeM pooling to local patch tokens. We experiment with both options. The VLAD aggregator~\cite{Arandjelovi2013AllAV,Jgou2010AggregatingLD} is not included because it requires unsupervised training with VPR-specific datasets whereas we attempt to make our method training-free.

\begin{figure*}[t]
    \centering
    \includegraphics[width=1\linewidth]{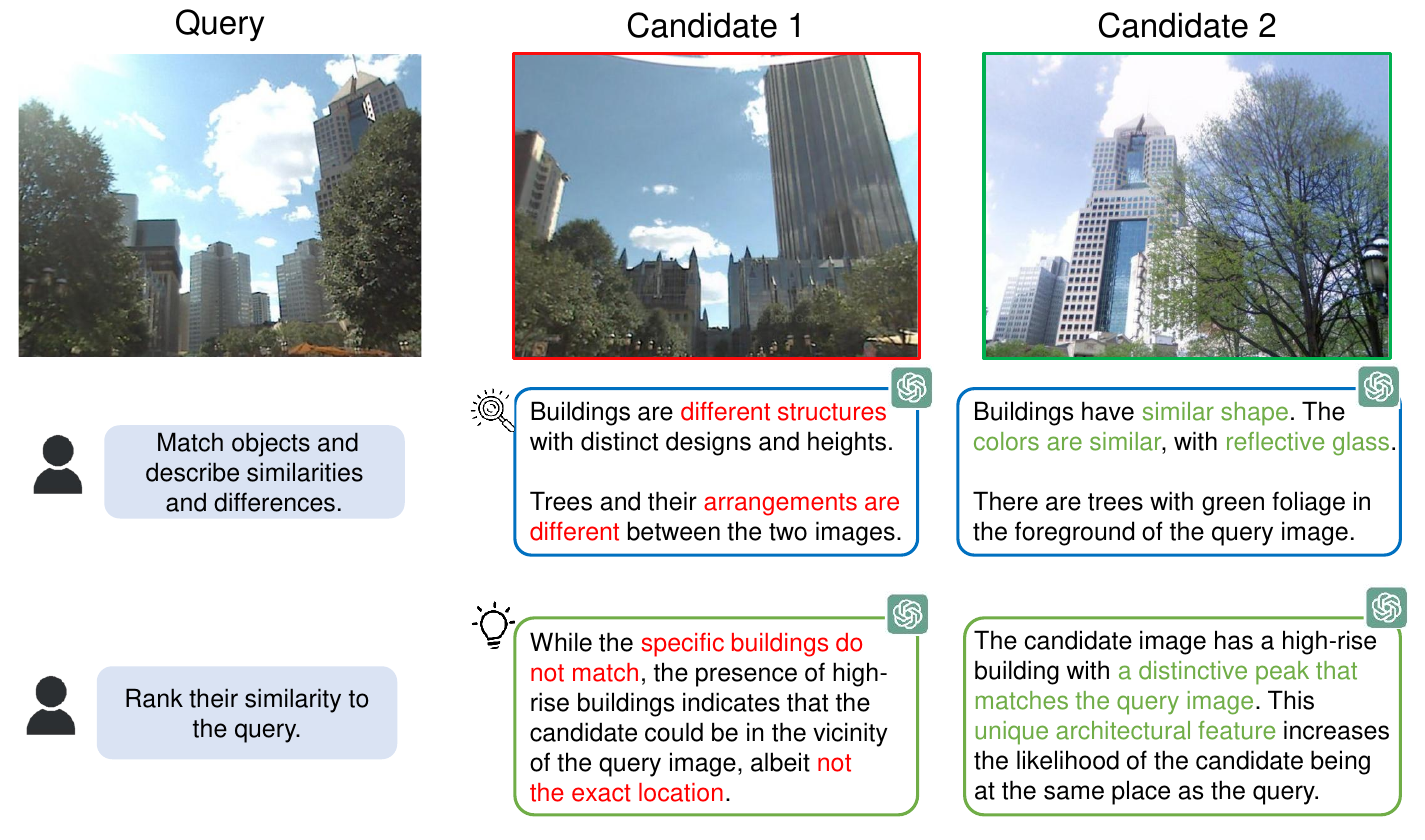}
    \caption{\textbf{Example of how our method works.} The query is selected from Pittsburgh30K~\cite{torii2013visual}, where candidate 1 is the top-1 retrieval of DINOv2 + GeM, and Candidate 2 is the Top-1 retrieval of our method. Correct retrieval is with \textcolor{green}{green} border, while the incorrect one is with \textcolor{red}{red} border.}
    \label{fig:demo}
\end{figure*}

\subsection{Vision-Language Refiner} After retrieving the top-K candidates, a vision-language refiner generates verbal descriptions for the delta of each query-candidate pair and reasons to finalize the ranking. 

\textbf{Choice of MLLMs.} The main challenge in leveraging verbal descriptions is to generate detailed descriptions relevant to VPR and filter out details irrelevant to VPR, such as weather and lighting. Unaware of the task-specific relevance of the visual details, the MLLMs need to be instructed on how to filter out irrelevant details and pay attention to the relevant ones via careful prompting. Meanwhile, detailed text descriptions of all retrieved candidates form a long context, and generating the final ranking based on them requires the MLLM to have sufficiently good long-text understanding capabilities.
In light of those, we choose GPT4-V~\cite{Achiam2023GPT4TR} for the vision-language refiner as it fulfills the above considerations. We expect similar performance from other comparable models.

\textbf{Design of Prompts.} As for the prompting, it is hard to verbally describe a scene such that humans can imagine all the details. However, telling the \textit{delta} between images is achievable and informative. So each query-candidate pair is sent to GPT-4V with a prompt asking for the similarities and differences between them, similar to the ``Bob and Alice'' scenario shown in Fig.~\ref{fig:teaser}. 
Additionally, we find it is important to clearly state what details should not be considered in the prompt and prevent the model from inferring imagined objects due to model hallucination. Moreover, it is informative to instruct GPT4-V that it needs to recognize the texts in the scene, just as Bob can recognize McDonald's because of the texts on its logos in the ``Bob and Alice'' example.

\textbf{Example Illustration.} One example from Pittsburgh30K is provided in Fig.~\ref{fig:demo} as an illustration of our method, where the shown prompt is a simplified version due to space limit. 
Complete prompts are included in our \textcolor{blue}{Appendix}.
To ensure that the final ranking of retrieved images aligns with human consensus, we emphasize again in the reranking prompt that we are doing the VPR task and what is important in this task, to filter out accidentally generated VPR-irrelevant details. 
By generating descriptions of \textit{delta} between image pairs and ranking similarity based on them, our method simulates the comparing-then-reasoning behavior of humans.

\begin{table}[t]
    \centering
    \small
    \caption{\textbf{Evaluation Results.} $R^2$Former indicates the reranking is not performed, where $^*$ and $\dagger$ indicate that Top 10 and Top 100 candidates are reranked respectively. Ours CLS and Ours GeM indicate that our method is applied on candidates retrieved with DINO CLS token and GeM pooled DINO features respectively. The best performances are in \textbf{bold} and the second best is underlined.}
    \label{tab:eval}
    \resizebox{1\textwidth}{!}{
    \begin{tabular}{c|c|cc|cc|cc}
        \Xhline{4\arrayrulewidth}
        \multirow{3}{*}{Category}&\multirow{3}{*}{Method} & \multicolumn{6}{c}{Dataset} \\
        \cline{3-8}
        & & \multicolumn{2}{|c|}{Tokyo247} & \multicolumn{2}{|c}{Baidu Mall} & \multicolumn{2}{|c}{Pittsburgh30K}\\
        \cline{3-8}
        
         & &{ R @ 1} & { R @ 5}  & { R @ 1} & { R @ 5} & { R @ 1} & { R @ 5}\\

        \hline
        \multirow{4}{*}{Supervised}&$R^2$Former & 65.1 & 79.4  & 53 & 71 &73.8 & 90.4  \\
        &$R^2$Former$^*$ & 80.3 & 85.1  & \underline{67.2} & \underline{75.5} & 85.9 & 92.8\\
        &$R^2$Former$\dagger$ & \textbf{88.6} &91.4  & \textbf{75.2} & \textbf{84.8} & \underline{87.6} & 94.9 \\
        &MixVPR& 76.8 &86.7 & 64.0 & 79.8 & \textbf{91.5} & \underline{96.7}\\
        \hline
        \multirow{2}{*}{Training Free}&DINO CLS & 61.3 & 81.9  & 41.5 & 51.2 
        &81.7 & 94.9 \\

        &DINO GeM & 81.9 & \underline{93.7} & 51.5 & 67 
        &86.4 & 96.4\\

        \hline
        \multirow{2}{*}{LLM-VPR}&Ours CLS & 73.0 (11.7$\uparrow$)  & 84.1 (2.2$\uparrow$) & 51.5 (10.0$\uparrow$) & 61.8 (10.6$\uparrow$) &83.0 (1.3$\uparrow$) & 95.3 (0.4$\uparrow$) \\
        &Ours GeM & \underline{87.0} (5.1$\uparrow$) & \textbf{94.3} (0.6$\uparrow$) & 62.0 (10.5$\uparrow$) & 71.5 (4.5$\uparrow$) &87.0 (0.6$\uparrow$) & \textbf{97.1} (0.7$\uparrow$)\\  
        \Xhline{4\arrayrulewidth}
    \end{tabular}
    }

\end{table}

\section{Experiments}

\textbf{Datasets.} Due to our budget constraints and the rate limit imposed in GPT-4 usage, we evaluate our method on Tokyo247~\cite{torii201524}, Baidu Mall~\cite{sun2017dataset}, and Pittsburgh30K~\cite{torii2013visual}. Tokyo247 and Baidu Mall contain non-English signage that is challenging for multimodal LLMs trained mainly with English data, while Pittsburg30K contains repetitive structures where the difference is hard to describe. Tokyo247 contains street view data collected in Tokyo, including both daytime and nighttime. There are 315 queries and around 76K references in this dataset. Baidu Mall is an indoor dataset collected at a shopping mall in China. There are 2293 queries and 689 references in this dataset, and we randomly subsample 400 queries for evaluation. Pittsburgh30K test split contains 6816 queries and 10000 references collected from driving vehicles in downtown Pittsburgh. We randomly subsample 1000 queries from this dataset. Though we only selected three datasets, they cover the majority of real-world daily scenarios: indoor, street views, and driving. Random subsampling is applied to only reduce the costs on GPT-4, where the data distribution is preserved.

\textbf{Evaluation Metrics.} We use Recall at K as our evaluation metric. If one of the top-K candidates is correct (geographical location is within $m$ meters of the query), the query is \textit{correctly recognized}. The Recall at K is the ratio of the number of correct retrievals and the total number of queries. The threshold m is set to 25 in the Tokyo247 and the Pittsburgh30K datasets and 10 in the Baidu Mall dataset, following~\cite{Zhu2023R2FU} and~\cite{AnyLoc} respectively.

\textbf{Method Setup.} Top-10 coarse retrievals are fed to our vision-language refiner. The output feature of the \texttt{[CLS]} token from DINOv2~\cite{oquab2023dinov2} and GeM aggregated descriptors of all patch tokens, denoted DINO CLS and DINO GeM, are selected as our coarse-retriever. We apply vision-language refiner to the top-10 candidates of these two methods, denoted \textit{Ours CLS} and \textit{Ours GeM}. 

\textbf{Baseline Methods}. We include $R^2$Former~\cite{Zhu2023R2FU} and MixVPR~\cite{Alibey2023MixVPRFM} as the supervised baselines because they achieve the SOTA performance among reranking methods and global retrieval methods. We implement $R^2$Former to rerank both Top-10 candidates (same as our setup) and Top-100 (original setup) candidates. The performance retrieved by DINO CLS and DINO GeM are also reported to evaluate whether the vision-language refiner can improve over the vision-based coarse retriever.

\subsection{ Quantitative Results}
\begin{figure*}[t]
    \centering
    \includegraphics[width=1\linewidth]{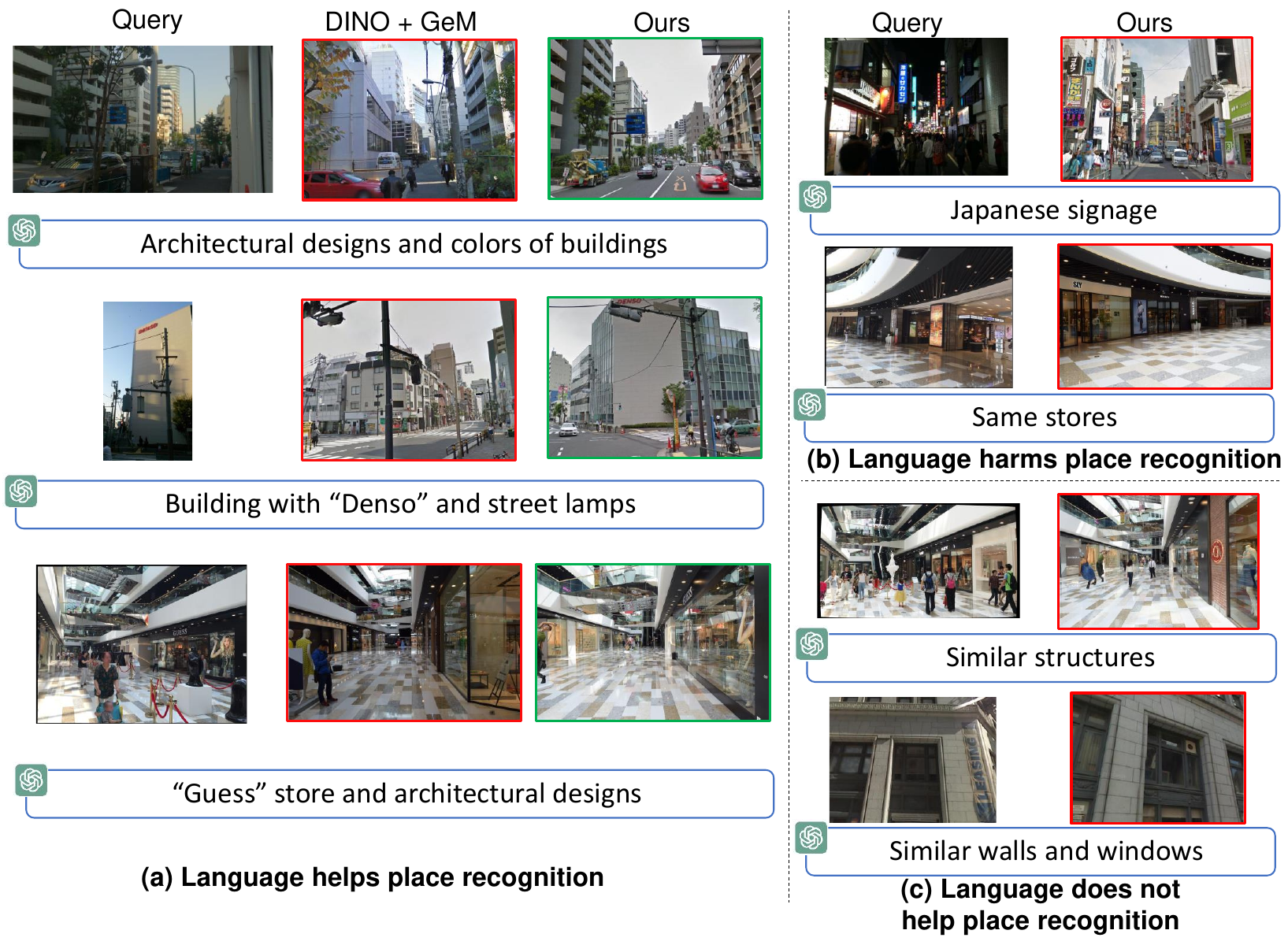}
    \caption{\textbf{Examples of (a) success cases, (b) fail cases, and (c) `cannot help' cases.} Correct retrievals (top 1) are in \textcolor{green}{green} border, and incorrect ones are in \textcolor{red}{red} border. The top two rows in (a) and the first row in (b) are selected from Tokyo247. The last row in (c) is selected from Pittsburgh30K. The other examples are selected from Baidu Mall. Text boxes are summarized GPT4-V outputs.\vspace{-3mm}}
    \label{fig:example}
\end{figure*}

The quantitative result is shown in Tab.~\ref{tab:eval}. Our vision-language refiner achieves comparable results with supervised baselines and steadily outperforms vision-only training-free baselines.
 
 \textbf{LLM-VPR largely improves over the coarse retriever.} In Baidu Mall, shopfronts are relatively easy to recognize since there are usually signages, resulting in satisfactory performance of the vision-language refiner. In Tokyo247, we can usually find unique landmarks to describe verbally to help localize the query. However, in Pittsburgh30K, there exists a large number of repetitive and similar structures, which makes it hard to verbally describe the similarities and differences. Even in this case, our vision-language refiner is still capable of improving VPR performance. 
 
 \textbf{LLM-VPR is comparable with supervised methods.} In Tokyo247 and Pittsburgh30K, the performance of our method is comparable to the best supervised baselines in R@1 and better than them in R@5. In Baidu Mall, our method is comparable with MixVPR and $R^2$Former that reranks top-10 candidates. In daily scenarios, the observed scenes can usually be described via language, where the semantic and informative information assists place recognition.
 
 \textbf{LLM-VPR does not achieve expected results in shopping malls.} LLM-VPR is outperformed by the $R^2$Former that reranks top-100 candidates with a big margin in Baidu Mall, but LLM-VPR is comparable with $R^2$Former in the other two datasets. Intuitively, the shopfronts are easy to describe and recognize, and we should expect a stronger performance of LLM-VPR. However, shopping malls can also introduce challenges. There might be stores with the same brand at different locations, which confuses the vision-language refiner. The repetitive structures in shopping malls, such as decorations, also introduce challenges when shopfronts do not clearly appear in the image. Moreover, multimodal LLMs are usually insensitive to the displacement of a camera when the camera looks at the same objects (details will be discussed in section~\ref{sec:qual}). Only candidates within 10 meters of the query will be considered correct in Baidu Mall, and this will amplify the negative impact of insensitivity to camera displacement.

\subsection{Qualitative Analysis}
\label{sec:qual}

\textbf{When do multimodal LLMs help?} It is interesting to investigate when and why multimodal LLMs help the VPR task. Some examples are shown in Fig.~\ref{fig:example}. In general, multimodal LLMs can help when there is \textit{(1) enough structural information that can be described, (2) unique landmarks, and (3) when MLLMs can avoid the perturbation from VPR-irrelevant items}. The camera viewpoint in the first row and bottom row in Fig.~\ref{fig:example} (a) contains enough structural information, so the structural similarities and differences such as textures, colors, or arrangements can be verbally described. If there are unique landmarks such as Fig.~\ref{fig:demo} and the last two rows in Fig.~\ref{fig:example} (a): building or signage, multimodal LLMs are able to recognize the place, even when the viewpoint is limited (the second query in Fig.~\ref{fig:example} (a)). English texts can help MLLMs as they can directly compare the contents, but if the texts are not in English, MLLMs tend to compare the shapes and colors. Due to the self-supervised nature of DINOv2~\cite{oquab2023dinov2}, the descriptor of an image may be negatively impacted by VPR-irrelevant items close to the camera. For example, in the top row in Fig.~\ref{fig:example} (a), there are street lamps close to the camera, covering part of the background. DINOv2 tends to pay attention to the street lamps, which is less important than other backgrounds in the VPR task. MLLMs can overcome such a problem by providing text descriptions and spatial reasoning.

\textbf{When do multimodal LLMs harm?} It is rare that MLLMs harm the performance, but there are some reasons why multimodal LLMs may harm place recognition: \textit{(1) there are a large number of items and (2) MLLMs are insensitive to the displacement of the camera when looking at the same item}. When images contain a large number of items, such as the first row in Fig.~\ref{fig:example} (b), it is too hard to similarities and differences for all items. Occasionally, MLLMs fail in some ``simple" examples because it is insensitive to the displacement of the camera when it looks at the same item. In the second row of Fig.~\ref{fig:example} (b), both images look at the same store (in Chinese texts), but we can see that their camera positions are relatively far in an indoor environment. When converting these two images into texts, such subtle changes in camera positions are hard to describe, and therefore MLLMs are not sensitive to the change in distance. As a result, MLLMs may harm VPR.

\textbf{When do multimodal LLMs \textit{fail} to help?} When the query-candidate pair contains \textit{repetitive and highly similar structures}, then MLLMs may not be able to tell the differences, such as the first row in Fig.~\ref{fig:example} (c). The shopping mall contains repetitive structures like walls and evaluators. Without a hint of nearby stores, it is hard to tell the differences between the two images. In addition, if the viewpoint is highly limited without any cues such as signage, similar to the last row in Fig.~\ref{fig:example} (c), MLLMs struggle due to the limited descriptive information available.

In addition to these qualitative analyses, we include qualitative ablation studies in the \textcolor{blue}{Appendix}.

\section{Conclusion}

We explore the underexamined area of LLMs in visual place recognition (VPR) inspired by human activities shown in the \textit{Bob and Alice} example. By integrating vision-based retrieval to propose several candidate locations and leveraging language-based reasoning to inspect and decide on the best candidate, we develop LLM-VPR, a robust VPR solution without any additional supervised training. Our method achieves comparable results with current supervised SOTA models, and language-based reasoning is shown to be able to successfully improve performance over vision-only baselines. We believe our work opens new possibilities for applying and designing foundation models, such as VFMs, LLMs, and MLLMs, to enhance the localization and navigation of mobile robots, thereby paving the way for more advanced and versatile robotic systems.

\clearpage
\acknowledgments{This work is supported by NSF Grant 2238968 and 2345139, and in part through the NYU IT High-Performance Computing resources, services, and staff expertise. We thank Gao Zhu for the valuable discussions at the beginning of this project.}

\bibliography{main}
\clearpage
\newpage
\appendix
\appendixpage

\startcontents[sections]
\printcontents[sections]{l}{1}{\setcounter{tocdepth}{2}}

\section{Ablation Studies} 
\begin{figure*}[t]
    \centering
    \includegraphics[width=1\linewidth]{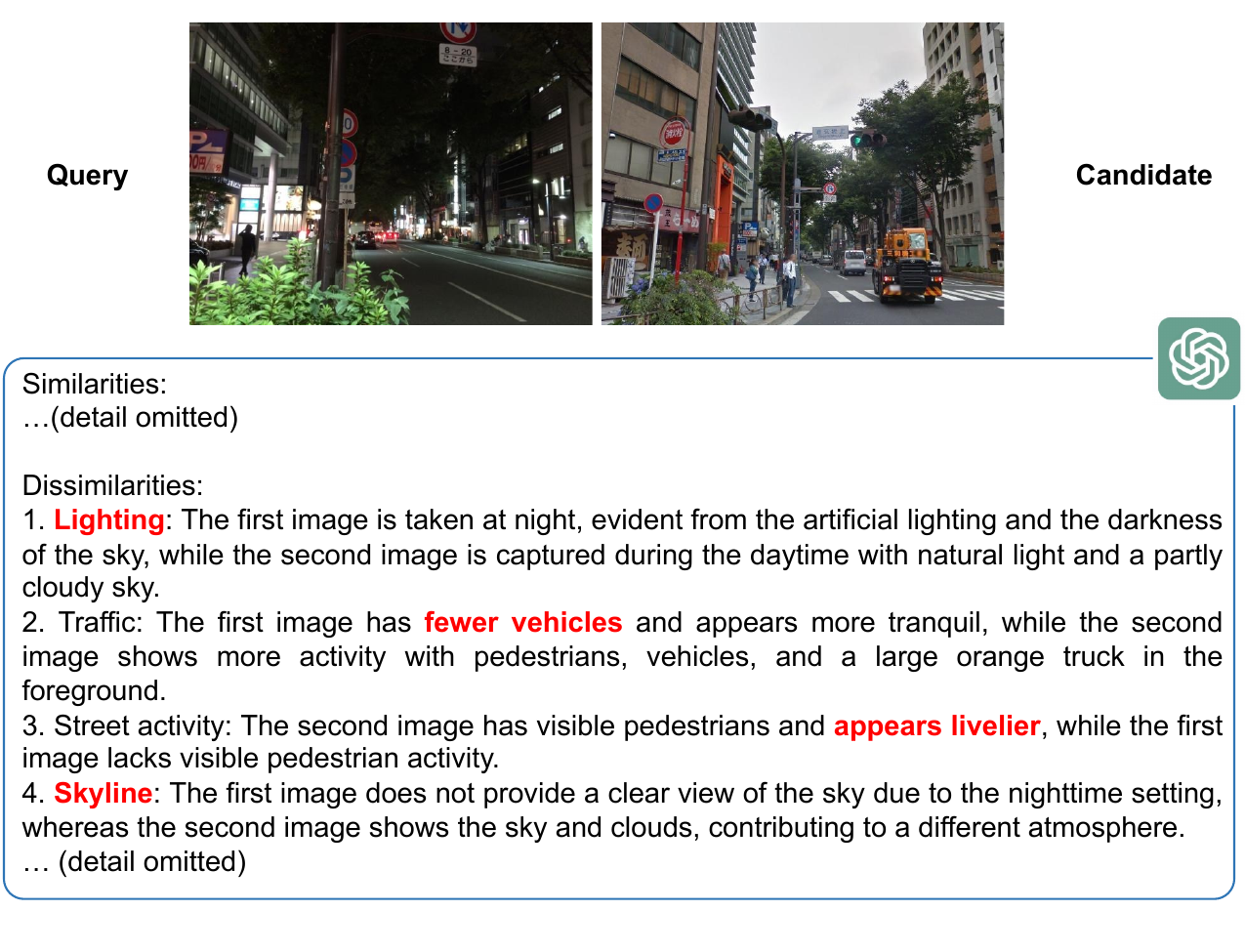}
    \caption{\textbf{Generated texts without constraints.} The image on the left is the query, and the image on the right is the candidate. Texts that may introduce irrelevant dissimilarities are in \textcolor{red}{red}.\vspace{-3mm}}
    
    \label{fig:text_nocon}
\end{figure*}

Our complete text generation prompt includes three main components: hints for describing similarities and dissimilarities,
hints for object matching, and constraints on irrelevant details and unnecessary inferences. The description of similarities and dissimilarities corresponds to the action of comparing in the ``Bob and Alice" example, while the other two are additional constraints. In Fig.~\ref{fig:text_noobj} and Fig.~\ref{fig:text_nocon}, we use one example from Tokyo247~\cite{torii201524} to showcase the influence of ablating the other two constraints. The ablated prompts are shown in Fig.~\ref{fig:prompt_nocon}.

\textbf{Hints on object matching.} We first discuss the impact of object matching. The prompt without object matching is in Fig.~\ref{fig:prompt_nocon} (a). The generated texts of the original prompt and prompt without object matching are shown in Fig.~\ref{fig:text_noobj}. With object matching, the logical flow of the text description becomes clear, vivid, and graphic. In contrast, when examining texts generated without object matching, it is more challenging to grasp the main points and form a clear mental image. Furthermore, with object matching, the similarities and differences are discussed in greater detail locally rather than broadly. 

\textbf{Constraints on irrelevant details.} If we further remove the constraints on irrelevant details and unnecessary inference (inferring unseen or unclear items), for which the prompt is shown in Fig.~\ref{fig:prompt_nocon} (b), VPR-irrelevant details such as lighting, vehicles, activities of a street, and skylines are mentioned, inducing noises during reasoning, as shown in Fig.~\ref{fig:text_nocon}. 

\section{Limitation and Future Research Direction}

\begin{figure*}[t]
    \centering
    \includegraphics[width=1\linewidth]{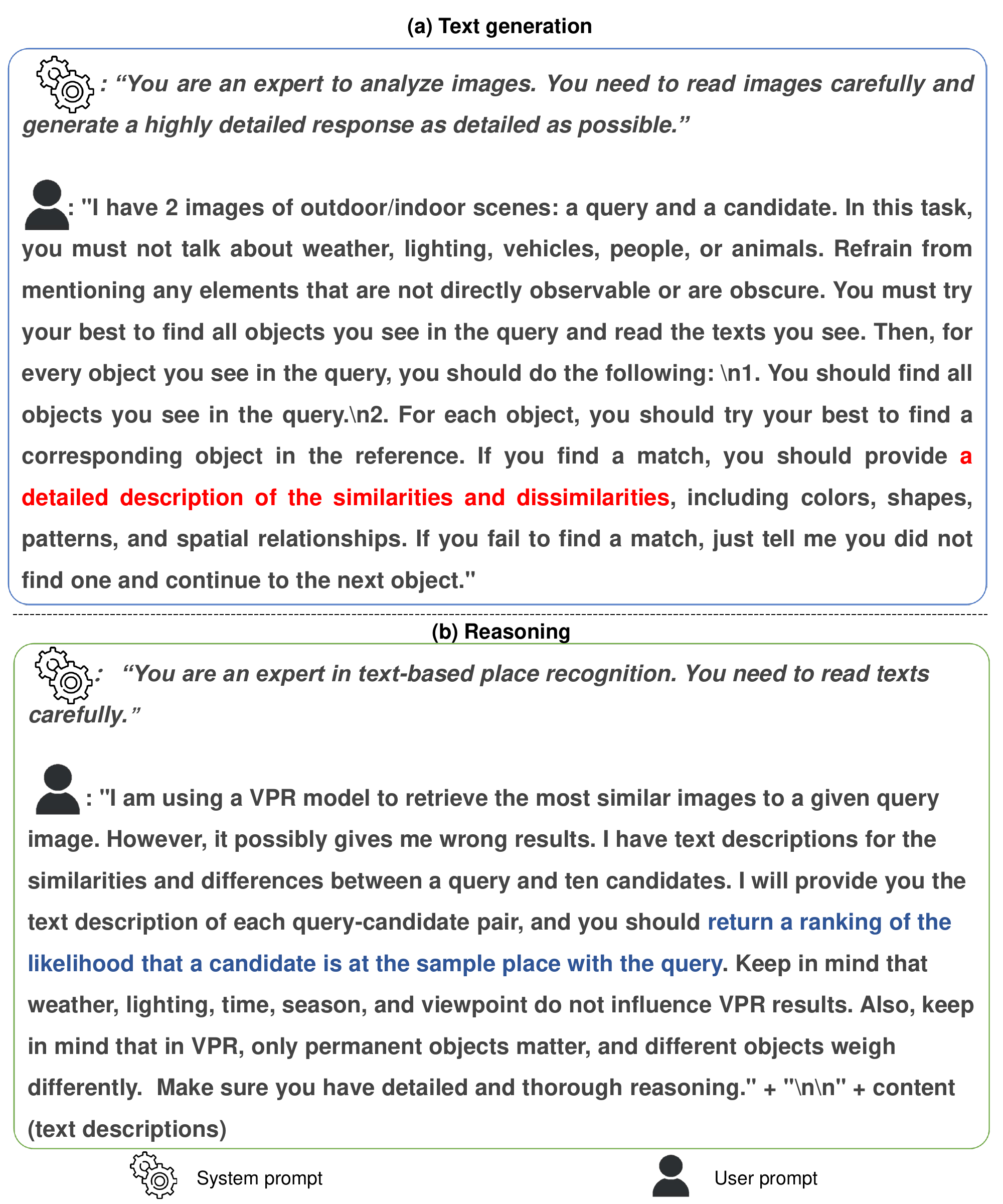}
    \caption{\textbf{(a) Text generation prompt and (b) Reasoning prompt.} Indoor or outdoor are selected based on different datasets. \vspace{-3mm}}
    \label{fig:prompt}
\end{figure*}

\begin{figure*}[t]
    \centering
    \includegraphics[width=1\linewidth]{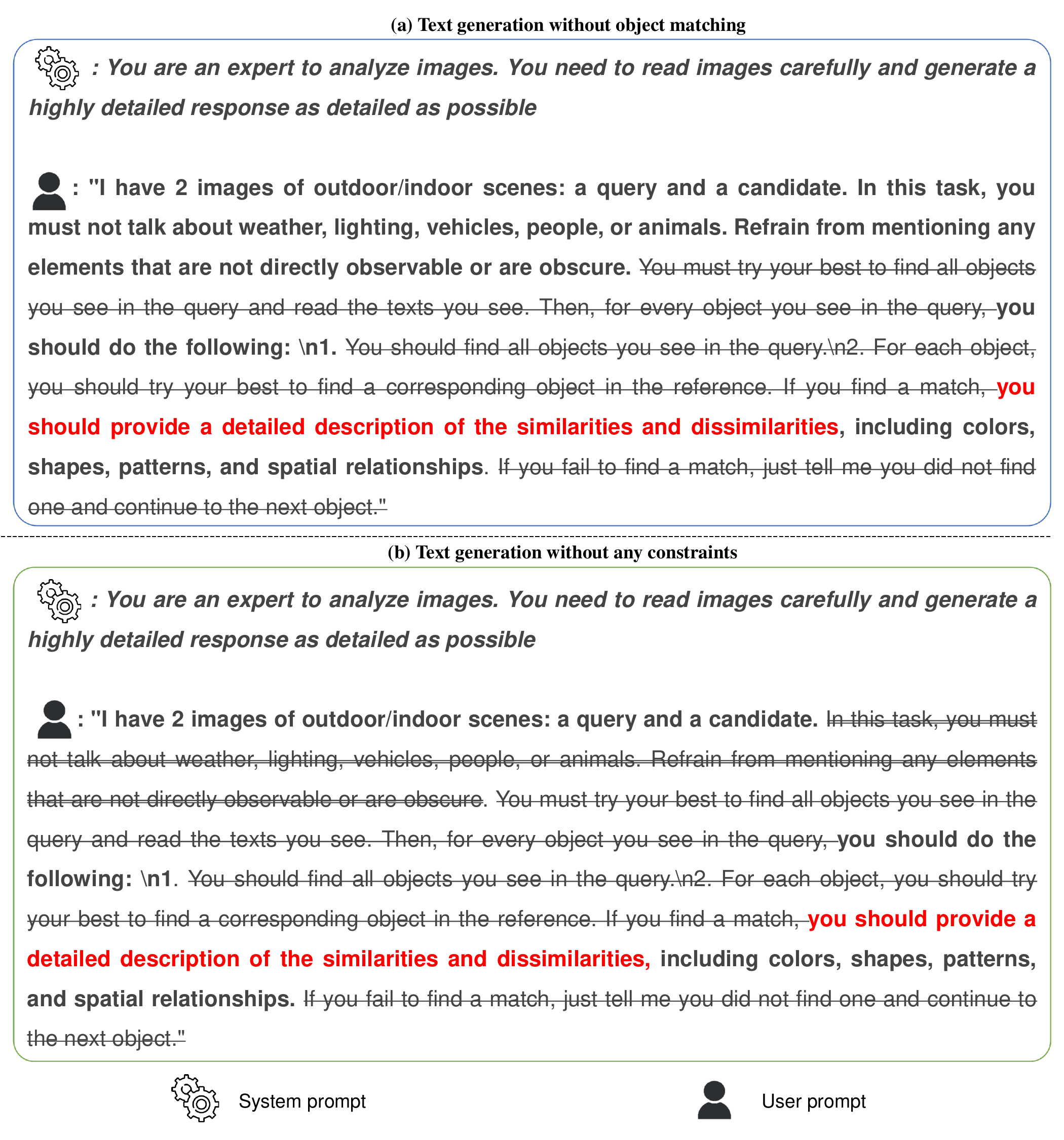}
    \caption{\textbf{Prompt ablation study.} The content about object matching is removed in (a), and the content about all constraints is removed in (b). Kept contents are in \textbf{bold}, the removed object matching contents are struck out, and removed constraints are doubly struck out.}
    \label{fig:prompt_nocon}
\end{figure*}

\begin{figure*}[t]
    \centering
    \includegraphics[width=\linewidth]{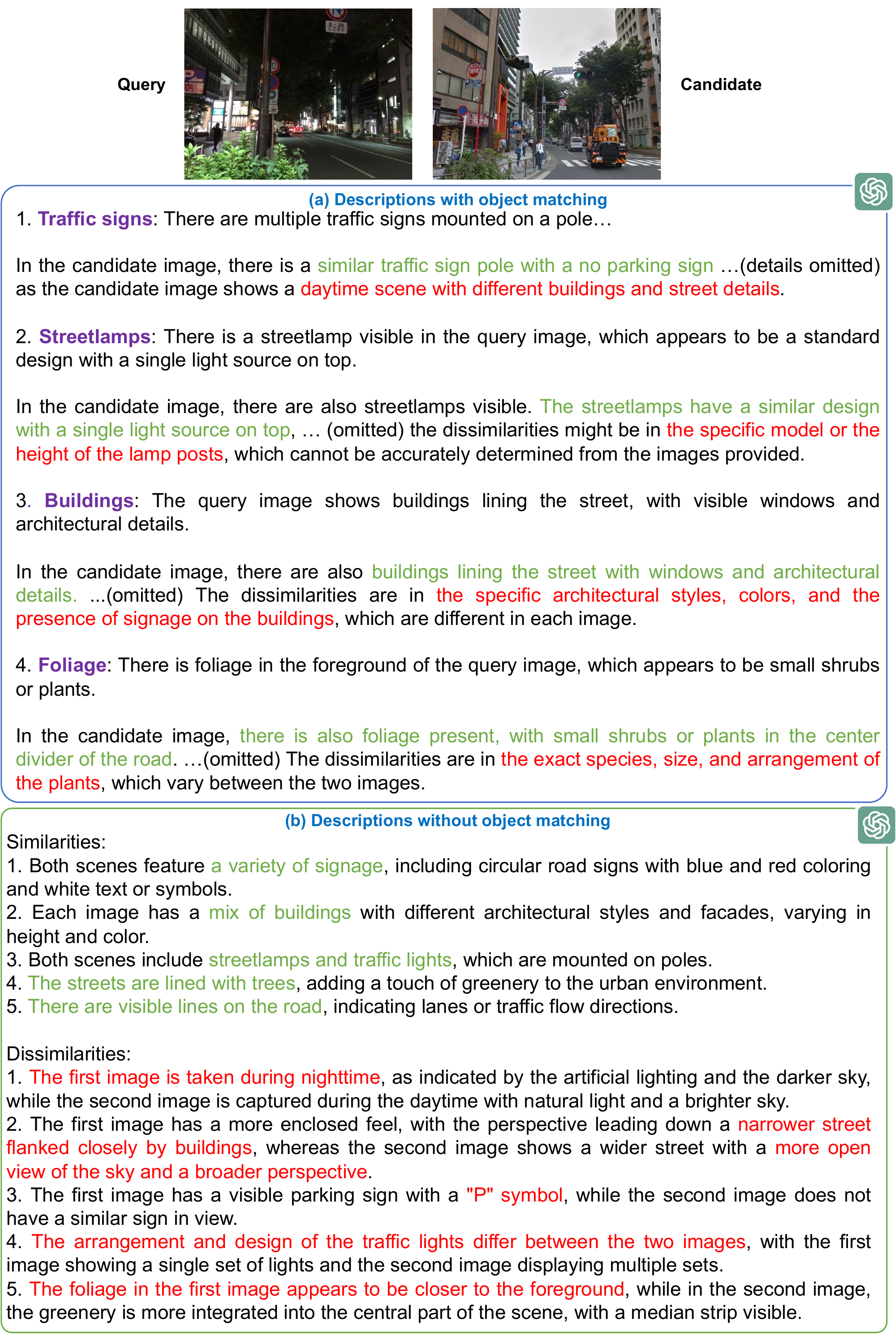}
    \caption{\textbf{Generated texts (a) with object matching and (b) without object matching.} \textcolor{green}{Green} texts show similarities, and \textcolor{red}{red} texts show differences. \textcolor{violet}{Violet} texts indcate matched objects.\vspace{-3mm}}
    
    \label{fig:text_noobj}
\end{figure*}

\begin{figure*}[t]
    \centering
    \includegraphics[width=\linewidth]{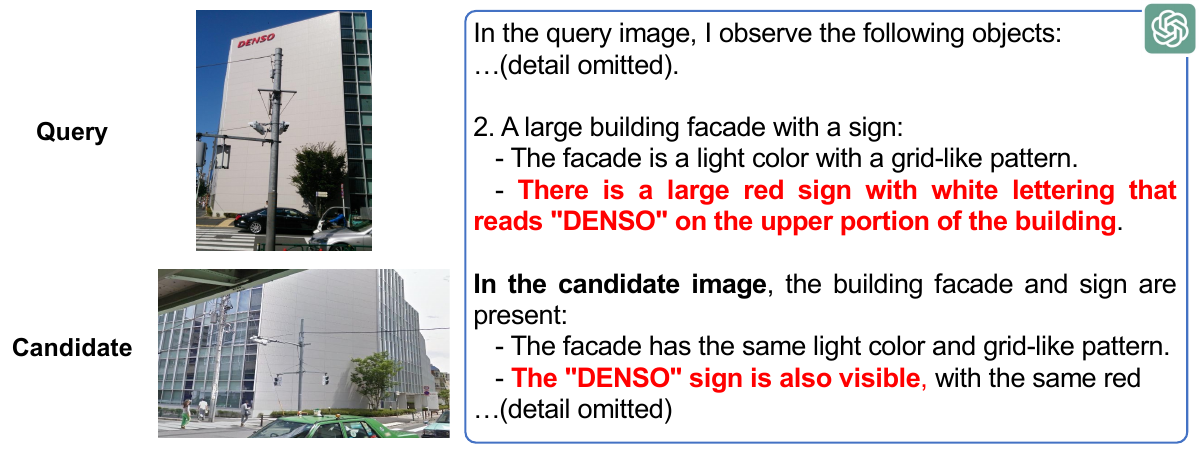}
    \caption{\textbf{Example of the hallucinated descriptions.} Non-factual descriptions due to hallucination are in \textcolor{red}{red}. The candidate does not contain the ``DENSO" text though the building looks similar to the query.\vspace{-3mm}}
    
    \label{fig:hallucination}
\end{figure*}

Although this work shows multimodal LLMs can be beneficial in place recognition and achieve good performance in a training-free pipeline, there are several limitations. 
\begin{enumerate}[leftmargin=1.3em]

    \item The multimodal LLMs may not strictly follow the prompt. As shown in Fig.~\ref{fig:text_noobj}, GPT still mentions lighting (Dissimilarities 1) while the prompt in Fig.~\ref{fig:prompt_nocon} (a) explicitly prevents it from doing so.
    \item The multimodal LLMs can generate non-factual descriptions due to their hallucination problem. In Fig.~\ref{fig:hallucination}, it is obvious that the candidate does not contain any text, but when looking at it along with the query, the model mistakenly concludes that the ``DENSO" text also appears in it.
    \item Long context issue. Since all descriptions are concatenated and fed into the refiner, it might get too long for efficient and correct place recognition if more candidates are involved. So the vision-language refiner relies on the vision-based coarse retriever to provide coarse filtering.
    \item The current pipeline requires computing and time budget and an internet connection to the multimodal LLMs APIs, which is not always ideal when deployment is considered.
\end{enumerate}

Besides addressing the limitations above, future research may consider automatic prompting for place recognition tasks. While the current prompts are human-designed, AI-generated prompts could potentially also be beneficial. Meanwhile, the proposed pipeline is entirely training-free. Though it is beyond our current budget, it would be interesting to see whether MLLM finetuned on VPR datasets can exhibit better spatial understanding and stronger place recognition performance.

\end{document}